\documentclass[sigconf]{acmart}

\usepackage{balance}
\usepackage{enumitem}
\usepackage{graphicx}
\usepackage{multirow}
\usepackage{makecell}
\usepackage{array}

\makeatletter
\newcommand{\thickhline}{%
    \noalign {\ifnum 0=`}\fi \hrule height 1pt
    \futurelet \reserved@a \@xhline
}
\newcolumntype{"}{@{\hskip\tabcolsep\vrule width 1pt\hskip\tabcolsep}}
\makeatother

\AtBeginDocument{%
  \providecommand\BibTeX{{%
    \normalfont B\kern-0.5em{\scshape i\kern-0.25em b}\kern-0.8em\TeX}}}

\setcopyright{acmcopyright}
\copyrightyear{2021}
\acmYear{2021}
\setcopyright{acmcopyright}
\acmConference[MM '21]{Proceedings of the 29th ACM International Conference on Multimedia}{October 20--24, 2021}{Virtual Event, China} 
\acmBooktitle{Proceedings of the 29th ACM International Conference on Multimedia (MM '21), October 20--24, 2021, Virtual Event, China}
\acmPrice{15.00}
\acmDOI{10.1145/3474085.3475375}
\acmISBN{978-1-4503-8651-7/21/10}

\acmSubmissionID{1229}

\begin{document}
\fancyhead{}
\title{UACANet: Uncertainty Augmented Context Attention for Polyp Segmentation}

\author{Taehun Kim}
\email{taehoon1018@postech.ac.kr}
\affiliation{%
  \institution{Pohang University of Science and Technology}
  \streetaddress{77, Cheongam-ro, Nam-gu}
  \city{Pohang-si}
  \state{Gyeongsangbuk-do}
  \country{Republic of Korea}
  \postcode{37655}
}

\author{Hyemin Lee}
\email{lhmin@postech.ac.kr}
\affiliation{%
  \institution{Pohang University of Science and Technology}
  \streetaddress{77, Cheongam-ro, Nam-gu}
  \city{Pohang-si}
  \state{Gyeongsangbuk-do}
  \country{Republic of Korea}
  \postcode{37655}
}

\author{Daijin Kim}
\email{dkim@postech.ac.kr}
\affiliation{%
  \institution{Pohang University of Science and Technology}
  \streetaddress{77, Cheongam-ro, Nam-gu}
  \city{Pohang-si}
  \state{Gyeongsangbuk-do}
  \country{Republic of Korea}
  \postcode{37655}
}
\renewcommand{\shortauthors}{Kim, et al.}

\begin{abstract}
  We propose Uncertainty Augmented Context Attention network (UACANet) for polyp segmentation which considers an uncertain area of the saliency map. We construct a modified version of U-Net shape network with additional encoder and decoder and compute a saliency map in each bottom-up stream prediction module and propagate to the next prediction module. In each prediction module, previously predicted saliency map is utilized to compute foreground, background and uncertain area map and we aggregate the feature map with three area maps for each representation. Then we compute the relation between each representation and each pixel in the feature map. We conduct experiments on five popular polyp segmentation benchmarks, Kvasir, CVC-ClinicDB, ETIS, CVC-ColonDB and CVC-300, and our method achieves \textit{state-of-the-art} performance. Especially, we  achieve 76.6\% mean Dice on ETIS dataset which is 13.8\% improvement compared to the previous \textit{state-of-the-art} method. Source code is publicly available at \url{https://github.com/plemeri/UACANet}
\end{abstract}



\maketitle

\section{Introduction}
Image segmentation is one of the fundamental and challenging topics in computer vision. It aims to classify each pixel from a given image. Recent studies adopt image segmentation to the specific domain such as salient object detection which focus on classifying each pixel whether it belongs to the most salient object or not. Similar to the salient object detection, one can apply their techniques to the medical purposes.

Medical image segmentation is widely used technique such as classifying each organ in the given tomography images like pancreas segmentation \cite{oktay2018attention}, detecting cells from the microscopy images \cite{ronneberger2015u}, or discriminating abnormal regions from normal regions from the body such as brain tumor \cite{haghighi2020learning} or polyp segmentation \cite{fan2020pranet}.

Polyps are an abnormal tissue growth from a surface of our body and can be found in colon, rectum, stomach or even throat. In most cases, polyps are benign, which means they are not indicating illness or maliciousness. However, because polyps are potentially cancerous, so we need a long-term diagnosis including the growth of their sizes or location and whether it became malignant or not. Thus, detecting polyps in the given colonoscopy image is beneficial for aiding early diagnosis of polyp related diseases.

Previous polyp segmentation networks usually adopt their methodologies from salient object detection (SOD) since they share the main interest, attend more on salient (polyp) region than surrounding scene. Current \textit{state-of-the-art} methods in SOD which shows decent performance are highly related to the edge guidance \cite{yang2017edge, su2019selectivity}. However, acquiring additional edge data is often expensive. Reverse attention \cite{chen2018reverse} suggest using reverse saliency map to obtain boundary cues, but since the boundary region is highly related to the ambiguous saliency score, saliency map without reverse operation already has such boundary information.

In this paper, we propose Uncertainty Augmented Context Attention network (UACANet), augmenting uncertain area with respect to the saliency map which is highly related to the boundary information. Our method computes the region with ambiguous saliency score and combine with a foreground and a background area for context attention module. On top of a modified version of U-Net \cite{ronneberger2015u} structure network with additional encoders and decoder, we aggregate the feature map based on three areas with weighted summation to obtain a representative context vectors for each area. We then compute the similarity between the context vector and the feature map. We validate our method with five famous polyp segmentation benchmarks, Kvasir, CVC-ClinicDB, ETIS, CVC-ColonDB and CVC-300 and achieve \textit{state-of-the-art} performance among previous methods. 

\section{Related Work}

\subsection{Semantic Segmentation}

Fully Convolutional Network (FCN) \cite{long2015fully} introduced a fully convolutional neural network architecture to efficiently train a model to classify each pixel. They also utilized a multi-scale scheme with aggregating low-level feature maps from the early stages in the neural network. Noh \textit{et al.} proposed deconvolution network \cite{noh2015learning} to compensate a spatial information degradation due to the pooling operation by leveraging transposed convolution and unpooling methods. Pyramid Spatial Pooling Network (PSPNet) utilized grid-wise pooling and comprise multiple sizes of the grid to deal with multi-scale objects. Deeplab \cite{chen2017rethinking} constructed multiple convolution layers with different dilation rates to diversify receptive fields within a single module to extract a multi-scale context information. Dual Attention Network \cite{fu2019dual} brought a self-attention mechanism to the semantic segmentation network and comprise a dual path network for spatial and channel dimensions which incorporates local features with their global dependencies. Object Contextual Representation (OCR) \cite{yuan2019object} expanded a non-local operation to consider semantic region by aggregating the representation of the pixels from each class and compute a similarity with each pixel from the feature map. 

\subsection{Salient Object Detection}

Salient object detection (SOD) resembles semantic segmentation but has its own criterion. Rather than predicting the entire region in the given image and classifying an object class label, SOD focuses on the object itself, identifying more important regions from the surrounding areas. Since salient object does not have any specific object class, even if the task might seem superficially easier than multi-class segmentation, it is much harder when it comes to the accurate object detection and assuming their priorities among other surrounding objects. Unlike semantic segmentation, it is hard to say that there is a strong baseline model with a high performing yet simple architecture in SOD. 

Current \textit{state-of-the-art} methods exploit the boundary region of objects for complementary information by multi-task learning strategy which can be incorporated to enhance the quality of saliency map. Edge Guidance Network (EGNet) \cite{yang2017edge} used a bottom-up stream for edge detection branch and side-out fusion strategy which aggregates each path from the top-down stream for salient object branch. To clarify the terms that we use as bottom-up stream means from high-level feature level to low-level feature level from the backbone network which is commonly used for segmentation or heat-map related tasks such as pose estimation. Also, side-out fusion strategy aggregates multiple feature maps from the backbone with concatenation or summation. Boundary-aware Network (BANet) \cite{su2019selectivity} oppositely used a side-out fusion for boundary branch and single-stream for object branch, but rather than regarding edge detection as a separate task, they combine edge result and object result to generate saliency map. These methods qualitatively showed competitive results consistently \cite{wang2021salient} which proves that edge guidance helps better representation of objects in terms of edge-preserving results. However, acquiring additional edge dataset is often expensive and even with image processing techniques such as Canny edge detection \cite{4767851}, it contains redundant edges which are usually unrelated to the object. 

Reverse attention \cite{chen2018reverse} explicitly multiply a reverse area of prediction in order to capture residual details for the saliency refinement. However, based on both their experimental results and our experiments, the performance with or without reverse attention wasn't very different. However, their suggestion provided an intuition that even without explicit edge guidance, we can access to the edge related context with saliency map. Based on this idea, we broaden a reverse saliency map with additional \textbf{\textit{uncertain area}}, an ambiguous area that saliency score is neither biased to the foreground nor the background. Also, they suggested an efficient yet practical network architecture. The decoder at the end of the backbone feature map produces a global saliency map, and in the bottom-up stream, only the saliency map from the previous level is used unlike U-Net shape networks concatenated feature maps. The final prediction thus only requires a low-level feature map and saliency map from the previous level as a guidance. While U-Net like networks predicts saliency map at the last bottom-up layer, proposed architecture from Reverse attention \cite{chen2018reverse} which computes saliency map at the intermediate level feature map and saliency map from the bottom-up stream only.

\begin{figure*}[]
  \centering
  \includegraphics[width=\textwidth]{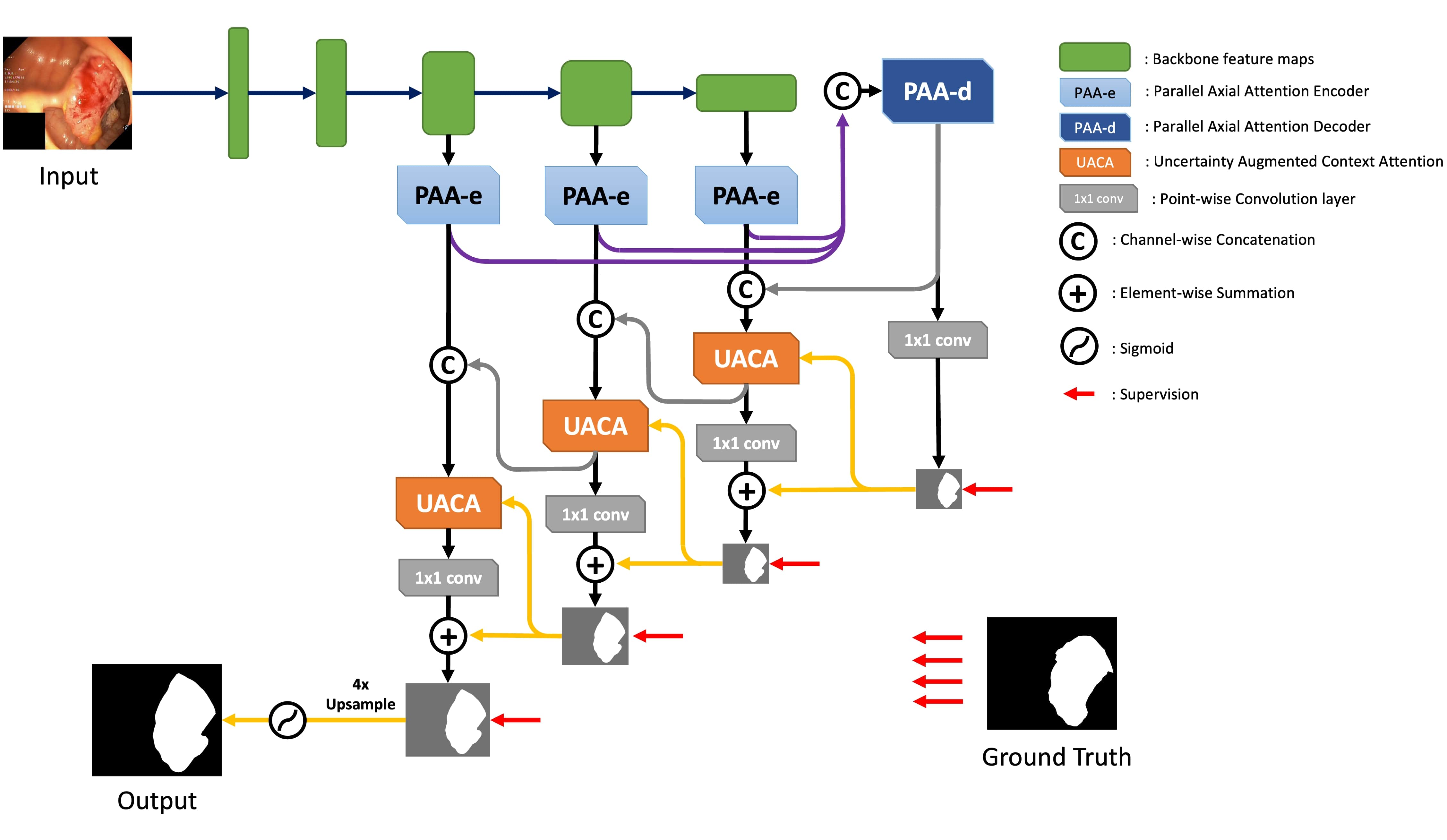}
  \caption{Overall architecture of UACANet.}
    \label{fig:1}
\end{figure*}

\subsection{Polyp Segmentation}

Polyp segmentation aims to precisely segment polyps from the given colonoscopy image. One can say that polyp segmentation is a semantic segmentation with binary class problem. While we can adopt such fully convolutional network architecture to solve the problem, since colonoscopy has different image domain compared to the general images, researchers focus on extracting semantic features with detail information. Famous architectures like PSPNet or Deeplab adopt multi-scale scheme to the backbone network which helps to capture detail information with multiple receptive field sizes, but they usually deploy such methods at the end of the backbone network which is spatially insufficient for recovering precise spatial information. DeeplabV3+ \cite{chen2017rethinking} tried to compensate such problem by concatenate low-level feature map from the backbone network but wasn't sufficient for recovering such details coming from the input image. 

U-Net \cite{ronneberger2015u} introduced incremental up-sampling of the feature maps alongside the corresponding scales of the low-level feature maps. Such "skip-connections" also appears in Feature Pyramid Networks (FPN) \cite{lin2017feature} but they use element-wise addition while U-Net aggregate features with channel-wise concatenation. While FPN is also designed similar to the U-Net in order to extract multi-scale features, they do not need a fine-grained detail features since object detection needs to locate a bounding box, not a precise shape of an object. U-Net++ \cite{zhou2018unet++} added extra layers and dense connectivity to reduce the gap between low-level and high-level features.

As the significance of polyp segmentation has been increased, studies that only dedicated to the polyp datasets have been done recently. ResUNet++ \cite{jha2019resunet++} construct a U-Net shape network with famous CNN modules including residual blocks from ResNet, Atrous Spatial Pyramid Pooling (ASPP) module from Deeplab and Squeeze and Excitation mechanism from SENet \cite{hu2018squeeze}. Selective Feature Aggregation network (SFA) \cite{10.1007/978-3-030-32239-7_34} added another bottom-up path for boundary estimation, similar to SOD networks with edge guidances \cite{yang2017edge, su2019selectivity}. Parallel Reverse Attention Network (PraNet) \cite{fan2020pranet} achieved \textit{state-of-the-art} performance on five different polyp segmentation benchmark by adopting the majority of the network design and techniques from \cite{chen2018reverse}. They added parallel partial decoder from \cite{wu2019cascaded} to combine low-level feature maps from the backbone network because the architecture from \cite{chen2018reverse} obviously lack of feature sharing for both top-down stream and bottom-up stream. In our network design, we locate intermediate encoder for low-level features and use the output of encoder and previous decoder features for bottom-up stream to incorporate high-level semantic features.

\section{Methodology}
\label{methods}

In this section, we demonstrate the architecture of our UACANet and the details of comprising modules. We first explain the overall structure of our network, then describe the details of fundamental components including Parallel Axial Attention encoder and decoder, and Uncertainty Augmented Context Attention.

\subsection{Overall Architecture}

We design UACANet based on the overall architecture from PraNet\cite{fan2020pranet} which is a modified version of \cite{chen2018reverse}. As shown in \figureautorefname~\ref{fig:1}, We add additional encoder network, Parallel Axial Attention encoder (PAA-e) for bottom-up stream and side-out fusion path. This helps to reduce the computational cost for both bottom-up and side-out fusion paths by reducing the number of channels for their input feature maps. Feature maps from three PAA-e modules are both used for side-out fusion path (purple arrow), Parallel Axial Attention decoder (PAA-d) and Uncertainty Augmented Context Attention (UACA). We concatenate three feature maps from PAA-e modules for PAA-d and it predicts the initial saliency map for polyps. Then the feature maps from PAA-e and PAA-d is concatenated for UACA, and the output saliency map from PAA-d is used for context guidance (yellow arrow). We describe detailed information of how UACA incorporates feature maps and context guidance in \sectionautorefname~\ref{subsec:3}. Then, the output saliency map from UACA is added with previously computed saliency map from the PAA-d. After the first UACA, we concatenate PAA-e feature map and previous UACA feature map for the next UACA (gray and black arrow with concatenation symbol). Also, the saliency map from the previous UACA is used as a context guidance for the next UACA (yellow arrows). The output of UACA is forwarded to the final point-wise convolution and then added with the previous context guidance for the current saliency map. After three consequent UACAs, the final output is computed with bi-linear up-sampling with scale factor of 4 and sigmoid function.

To sum up, the overall architecture shows that the backbone features are encoded with PAA-e, and encoded features are forwarded to PAA-d for initial saliency map which serves as a initial guidance map, which leads UACA to learn a residual saliency map apart from the initial map. This helps the consequent UACA can focus more on uncertain area like boundaries rather than fairly evident region.

We use both binary cross entropy (BCE) loss and intersection over union (IoU) loss. The loss function $\mathcal{L}$ is computed as follows,

\begin{equation}
    \begin{gathered}
    \label{eq:1}
    \mathcal{L}_{BCE} = -\sum_{i \in \mathcal{I}}{y_i log(\hat{y}_i) + (1-y_i)log(1-\hat{y}_i)}, \\
    \mathcal{L}_{IoU} = 1 - \frac{\sum_{i \in \mathcal{I}}y_i\hat{y}_i}{\sum_{i \in \mathcal{I}}y_i + \hat{y}_i - y_i\hat{y}_i}, \\ \\
    \mathcal{L} = \mathcal{L}_{BCE} + \mathcal{L}_{IoU},
    \end{gathered}
\end{equation}
where $i \in \mathcal{I}$ refers to a pixel in the output and ground truth, $y$ denotes ground truth and $\hat{y}$ denotes the output. As shown in \figureautorefname~\ref{fig:1}, we add four losses from four prediction from PAA-d and UACA with loss function in \equationautorefname~\ref{eq:1} (red arrows in \figureautorefname~\ref{fig:1}).
\begin{figure}[]
  \centering
  \includegraphics[width=\linewidth]{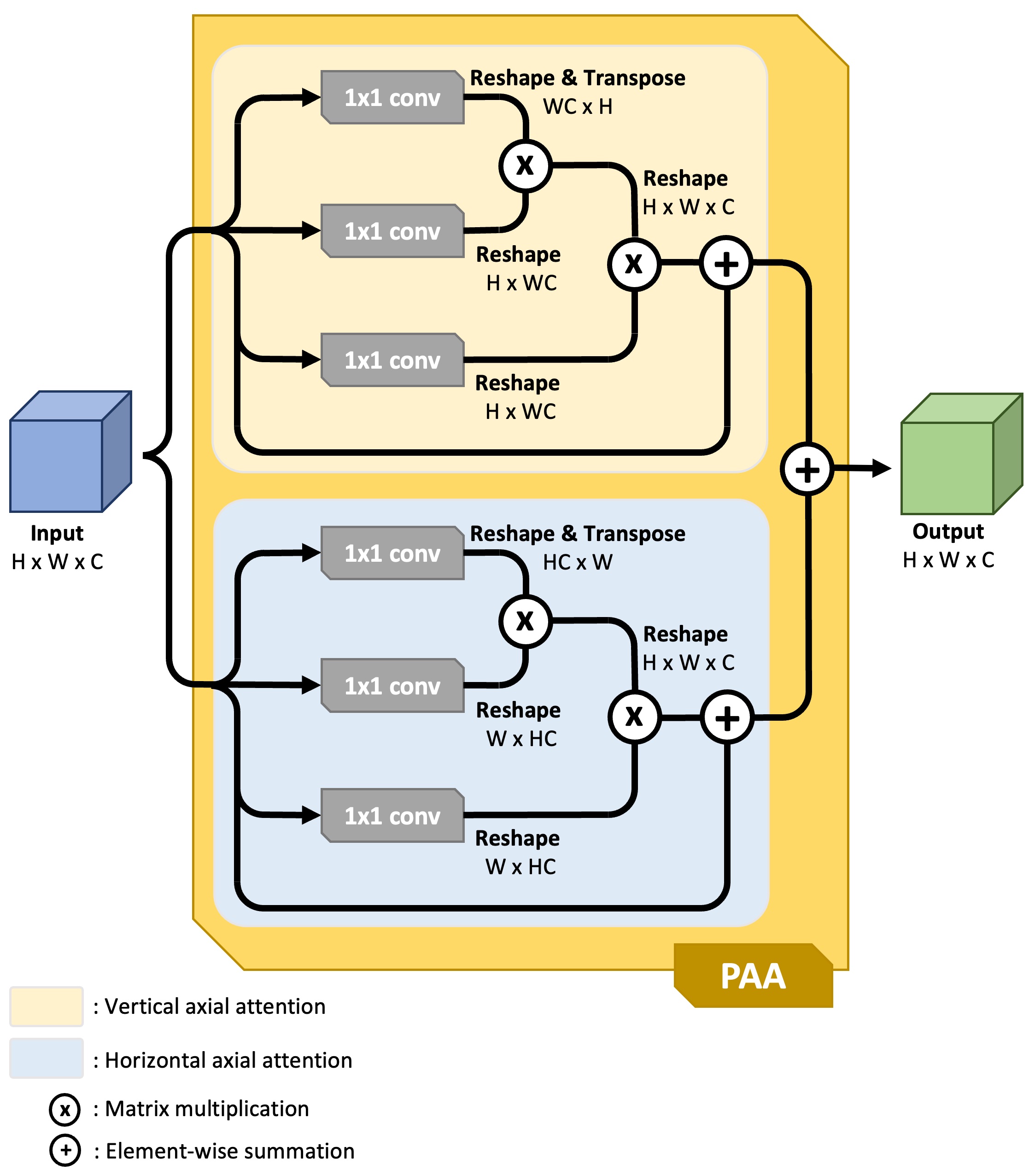}
  \caption{The details of Parallel Axial Attention (PAA)}
    \label{fig:2}
\end{figure}

\subsection{Parallel Axial Attention encoder and decoder}

In the era of deep learning, computer vision researchers who study on semantic segmentation and its related tasks like polyp segmentation are dedicated to find a better architecture to extract a fine-grained feature maps which have both high-level semantic information and low-level detail information because usually they are both hard to extract and to combine each other. Self-attention mechanism \cite{zhang2019self} is one of the major context modules in computer vision but requires heavy computation. Axial attention \cite{ho2019axial} solve this problem with performing non-local operation with respect to the single axis and sequentially connected each operation. 

We propose Parallel Axial Attention (PAA) for extracting both global dependencies and local representation. We adopt axial attention strategy by computing non-local operation for both horizontal axis and vertical axis but collocated in parallel. By locating vertical and horizontal attention, both contributes to the final output almost equally compared to the sequential method. While sequentially connected axial attention added trainable positional encoding, we do not use encoding scheme since positional encoding substantially not effective in relatively small scales. Also, we discover that using a parallel connection, element-wise summation is effective to aggregate feature maps rather than concatenation without performance degradation since the same input is used for both horizontal axis and vertical axis and they contributes to the output almost equally by the parallel connection. Also, since a single axis based attention cause unexpected deformation, element-wise summation can help to compensate such artifact. As shown in \figureautorefname~\ref{fig:2}, we compute two non-local operations with input feature map, one for horizontal axis and the other for vertical axis.

\begin{figure}[]
  \centering
  \includegraphics[width=\linewidth]{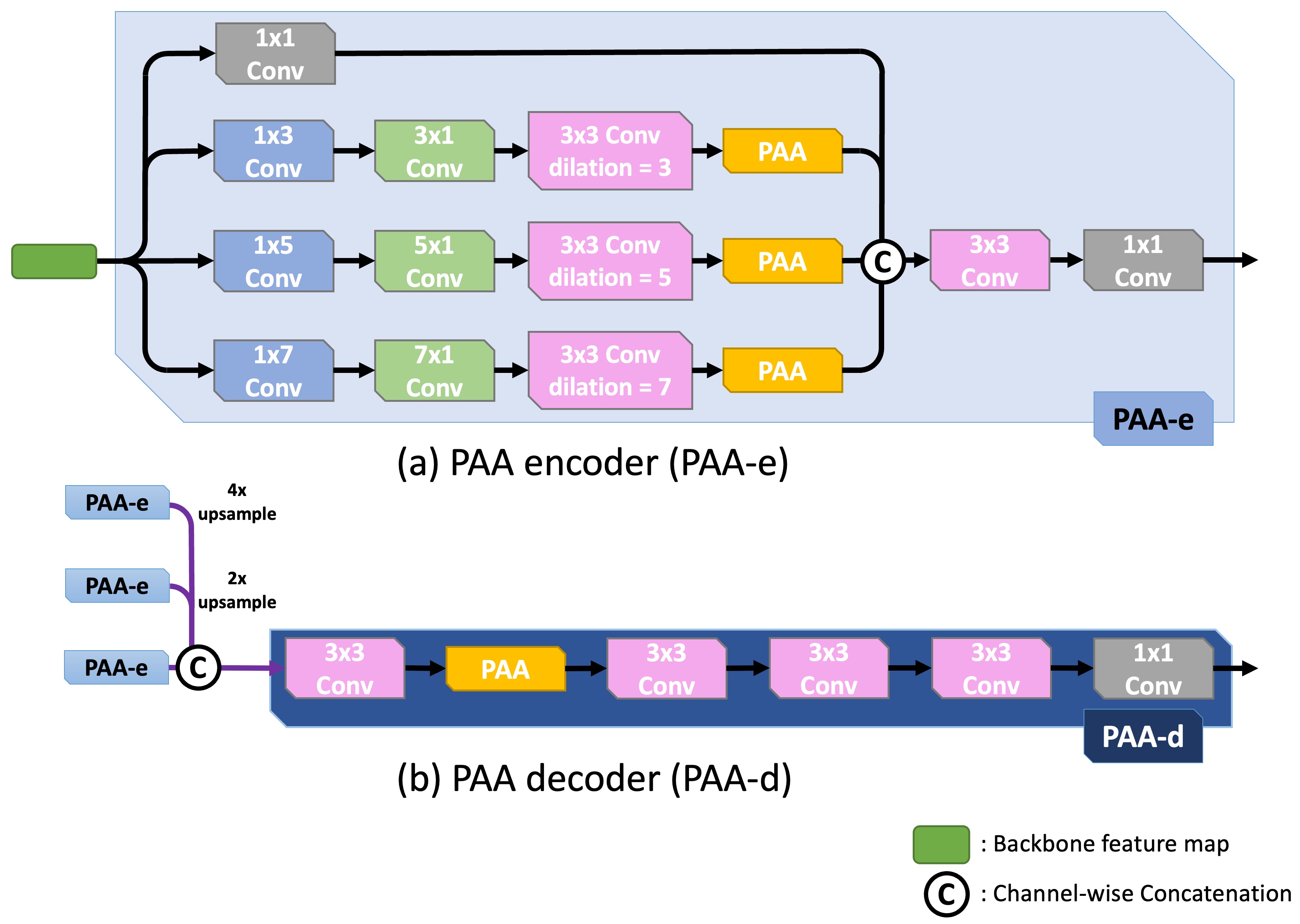}
  \caption{The details of Parallel Axial Attention encoder (PAA-e) (a) and decoder (PAA-d) (b)}
  \label{fig:3}
\end{figure}

We choose to actively use this module for encoder and decoder modules for better representation in terms of globally refined feature. First, we design  Parallel Axial Attention encoder (PAA-e) which aggregates the low-level feature maps from the top-down stream which will be used for bottom-up stream. Since U-Net structure use the low-level features without channel reduction, redundant information may hinder the performance and the number of channels are quite large since backbone networks are trained for image classification. Not to lose any detail information while reducing the number of channels, we design PAA-e with Receptive Field Block (RFB) \cite{liu2018receptive} strategy. As shown in \figureautorefname~\ref{fig:3}(a), feature map from the backbone network (green box) is forwarded to each receptive field path. We add PAA for additional global refinement for each scale and concatenate the outputs, then forward to the consecutive convolution layers. As shown in \figureautorefname~\ref{fig:1}, the outputs of PAA-e are used for both decoder module and bottom-up stream. Also, we design Parallel Axial Attention decoder (PAA-d) with simple structure yet add additional PAA for final feature aggregation from different level of PAA-e features which are denoted as purple arrows in \figureautorefname~\ref{fig:1} and \figureautorefname~\ref{fig:3}(b).

\subsection{Uncertainty Augmented Context Attention}

\label{subsec:3}

While reverse attention \cite{chen2018reverse} in both SOD and polyp segmentation does not bring the large margin of performance gain, it was evident that it has shown some qualitatively better results. This phenomenon is highly related to the boundary guided SOD networks \cite{yang2017edge, su2019selectivity},  which show \textit{state-of-the-art} performances in multiple SOD benchmarks. Boundaries of the object as a extra supervision in SOD networks tends to compensate false negative areas, in other words, object regions with low saliency scores. Reverse attention is thus potentially effective method to bring implicit edge guidance without explicit shape of boundary supervision.

\begin{figure}[]
  \centering
  \includegraphics[width=\linewidth]{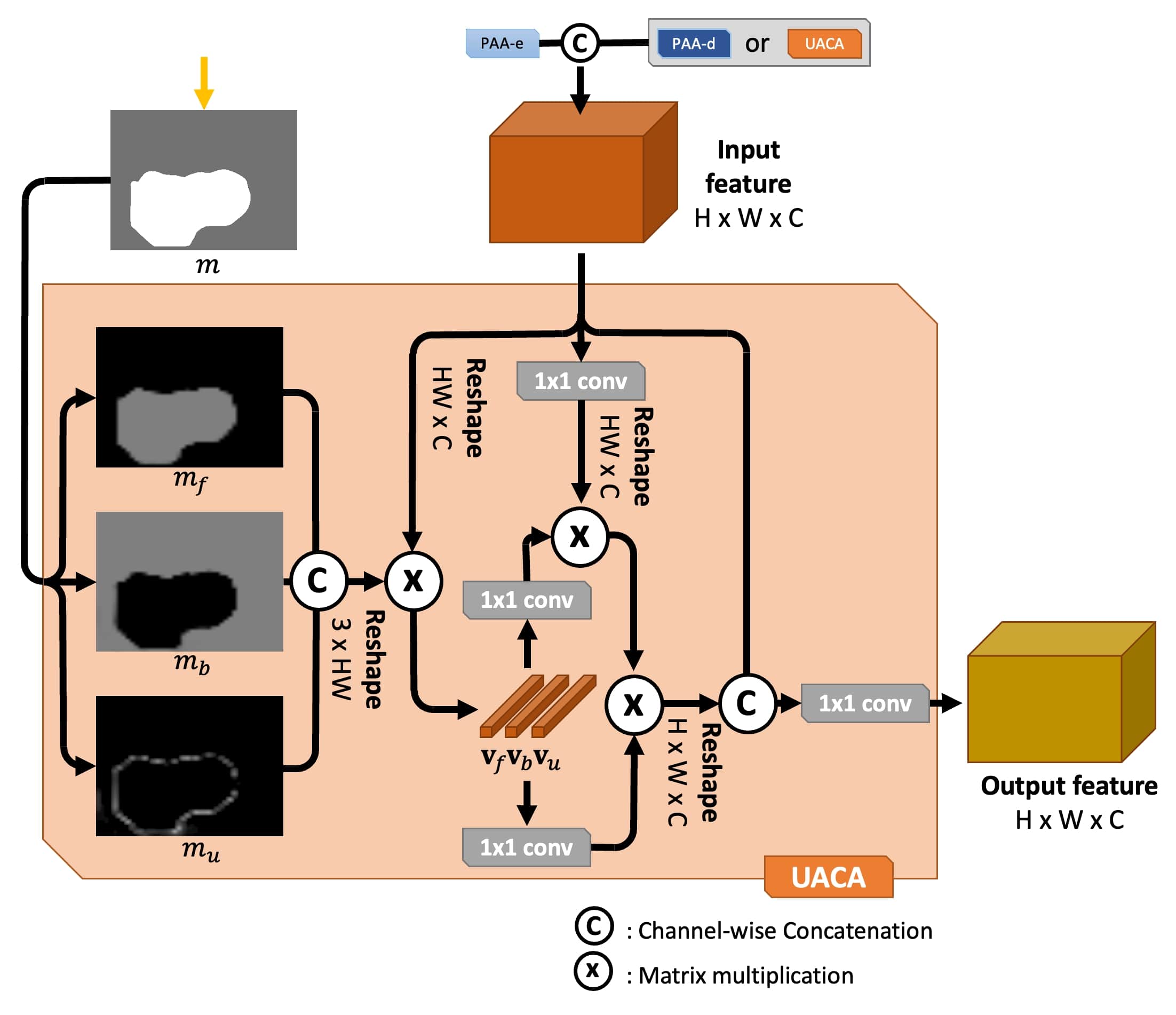}
  \caption{The details of Uncertainty Augmented Context Attention (UACA) module.}
    \label{fig:4}
\end{figure}

We focus on both saliency and reverse saliency map from the reverse attention and found that usually the boundary region appears where saliency score shows ambiguity. In other words, boundary region is highly related to the saliency score around $0.5$. From this property of saliency map, we assume that both saliency and reverse saliency map has almost equal amount of edge information since reverse saliency map is obtained by simple subtraction from 1. Based on this assumption, extracting ambiguous region from the saliency map as well as foreground and background area would improve attentive methods such as self-attention.

Based on our research, we propose Uncertainty Augmented Context Attention (UACA) module, a novel self-attention mechanism which incorporates \textbf{\textit{uncertain area}} for rich semantic feature extraction without extra boundary guidance. We denote previously computed input  saliency map as $\textbf{m}$ and generate corresponding foreground map $\textbf{m}_f$, background map $\textbf{m}_b$ and uncertain area map $\textbf{m}_u$ as follows,

\begin{equation}
    \label{eq:2}
    \begin{gathered}
    \textbf{m}_f = \text{max}(\textbf{m} - 0.5, 0), \quad
    \textbf{m}_b = \text{max}(0.5 - \textbf{m}, 0), \\
    \textbf{m}_u = 0.5 - \text{abs}(\textbf{m} - 0.5).
    \end{gathered}
\end{equation}
We compute foreground and background map with max operation in order to disentangle not only from each other but from the uncertain area map  since uncertain area map already represents their joint region which makes a redundant information and may diminish the role of uncertainty.

While reverse attention applies explicit channel-wise multiplication to the feature map which resembles Convolutional Block Attention Module (CBAM) \cite{woo2018cbam}, we compose our context module with a set of non-local operations. Similar to OCR \cite{yuan2019object} we first compute representative vectors for foreground map, background map and \textbf{\textit{uncertain area}} map by aggregating the pixel representation with each area map from the input feature map $\textbf{x}$ as follows,

\begin{equation}
    \label{eq:3}
    \begin{gathered}
    \textbf{v}_f = \sum_{i \in \mathcal{I}} \textbf{m}_{fi} \textbf{x}_i, \quad
    \textbf{v}_b = \sum_{i \in \mathcal{I}} \textbf{m}_{bi} \textbf{x}_i, \quad
    \textbf{v}_u = \sum_{i \in \mathcal{I}} \textbf{m}_{ui} \textbf{x}_i,
    \end{gathered}
\end{equation}
where $i \in \mathcal{I}$ denotes pixels in spatial dimension. We implement \equationautorefname~\ref{eq:2} with matrix multiplication as shown in \figureautorefname~\ref{fig:4}. Each vector stands for the representative feature vector, so $\textbf{v}_f$ represents foreground feature and $\textbf{v}_u$ represents the uncertain area. Then we compute the similarity between each representation vector ($\textbf{v}_f$, $\textbf{v}_b$ and $\textbf{v}_u$) and each pixel from the input feature map $\textbf{x}_i$ as follows,

\begin{equation}
\label{eq:4}
\begin{gathered}
s_{fi}' = \psi(\textbf{x}_i)^\top \phi(\textbf{v}_{f}), \quad
s_{bi}' = \psi(\textbf{x}_i)^\top \phi(\textbf{v}_{b}), \quad
s_{ui}' = \psi(\textbf{x}_i)^\top \phi(\textbf{v}_{u}), \\
s_{fi} = \frac{e^{s_{fi}'}}{N}, \quad
s_{bi} = \frac{e^{s_{bi}'}}{N}, \quad
s_{ui} = \frac{e^{s_{ui}'}}{N}, \quad \\
\text{where}, \quad \textit{N} = e^{\textbf{s}_{fi}'} + e^{\textbf{s}_{bi}'} + e^{\textbf{s}_{ui}'}.
\end{gathered}
\end{equation}

\begin{table}[]
    \begin{center}
        \resizebox{\linewidth}{!}{
        \begin{tabular}{ll|ccc}
        \thickhline
        Dataset & Method & Mean Dice $\uparrow$ & Mean IoU $\uparrow$ & MAE $\downarrow$ \\ \hline \hline
        \multirow{2}{*}{\makecell{CVC\\-ClinicDB}} & UACANet-S (w/o PAA) & 0.902 & 0.858 & 0.008 \\
        & UACANet-S & \textbf{0.916} & \textbf{0.870} & \textbf{0.008} \\ \thickhline
        \multirow{2}{*}{ETIS} & UACANet-S (w/o PAA) & 0.684 & 0.603& 0.029 \\
        & UACANet-S & \textbf{0.694} & \textbf{0.615} & \textbf{0.023}  \\ \thickhline
        \end{tabular}}
    \end{center}
    \caption{Ablation study for Parallel Axial Attention (PAA) on CVC-ClinicDB and ETIS datasets. $\uparrow$ denotes higher the better and $\downarrow$ denotes lower the better.}
    \label{tab:1}
\end{table}

\begin{table}[]
    \begin{center}
        \resizebox{\linewidth}{!}{
        \begin{tabular}{ll|ccc}
            \thickhline
            Dataset & Method & Mean Dice $\uparrow$ & Mean IoU $\uparrow$ & MAE $\downarrow$ \\ \hline \hline
            \multirow{4}{*}{\makecell{CVC\\-ClinicDB}} & CANet-S & 0.911 & 0.857 & 0.009 \\
            & CANet-L & 0.912 & 0.861 & 0.009 \\
            & UACANet-S & \textbf{\textcolor{blue}{0.916}} & \textbf{\textcolor{blue}{0.870}} & \textbf{\textcolor{blue}{0.008}} \\
            & UACANet-L & \textbf{\textcolor{red}{0.926}} & \textbf{\textcolor{red}{0.880}} & \textbf{\textcolor{red}{0.006}} \\ \thickhline
              \multirow{4}{*}{ETIS}    & CANet-S & 0.691 & 0.613 & 0.026 \\
            & CANet-L & 0.678 & 0.604 & \textbf{\textcolor{blue}{0.019}} \\
            & UACANet-S & \textbf{\textcolor{blue}{0.694}} & \textbf{\textcolor{blue}{0.615}} & 0.023  \\
            & UACANet-L & \textbf{\textcolor{red}{0.766}} & \textbf{\textcolor{red}{0.689}} & \textbf{\textcolor{red}{0.012}} \\ \thickhline
        \end{tabular}}
    \end{center}
    \caption{Ablation study for \textbf{\textit{uncertain area}} on CVC-ClinicDB and ETIS datasets. Red color denotes the best score among the methods, and blue color denotes the second best. $\uparrow$ denotes higher the better and $\downarrow$ denotes lower the better.}
    \label{tab:2}
\end{table}

Finally, we compute context feature map by weighted summation of representation vector $\textbf{v}_f$, $\textbf{v}_b$ and $\textbf{v}_u$ by similarity score $s_{f}$, $s_{b}$ and $s_{c}$ as follows,

\begin{figure*}[]
  \centering
  \includegraphics[width=\textwidth]{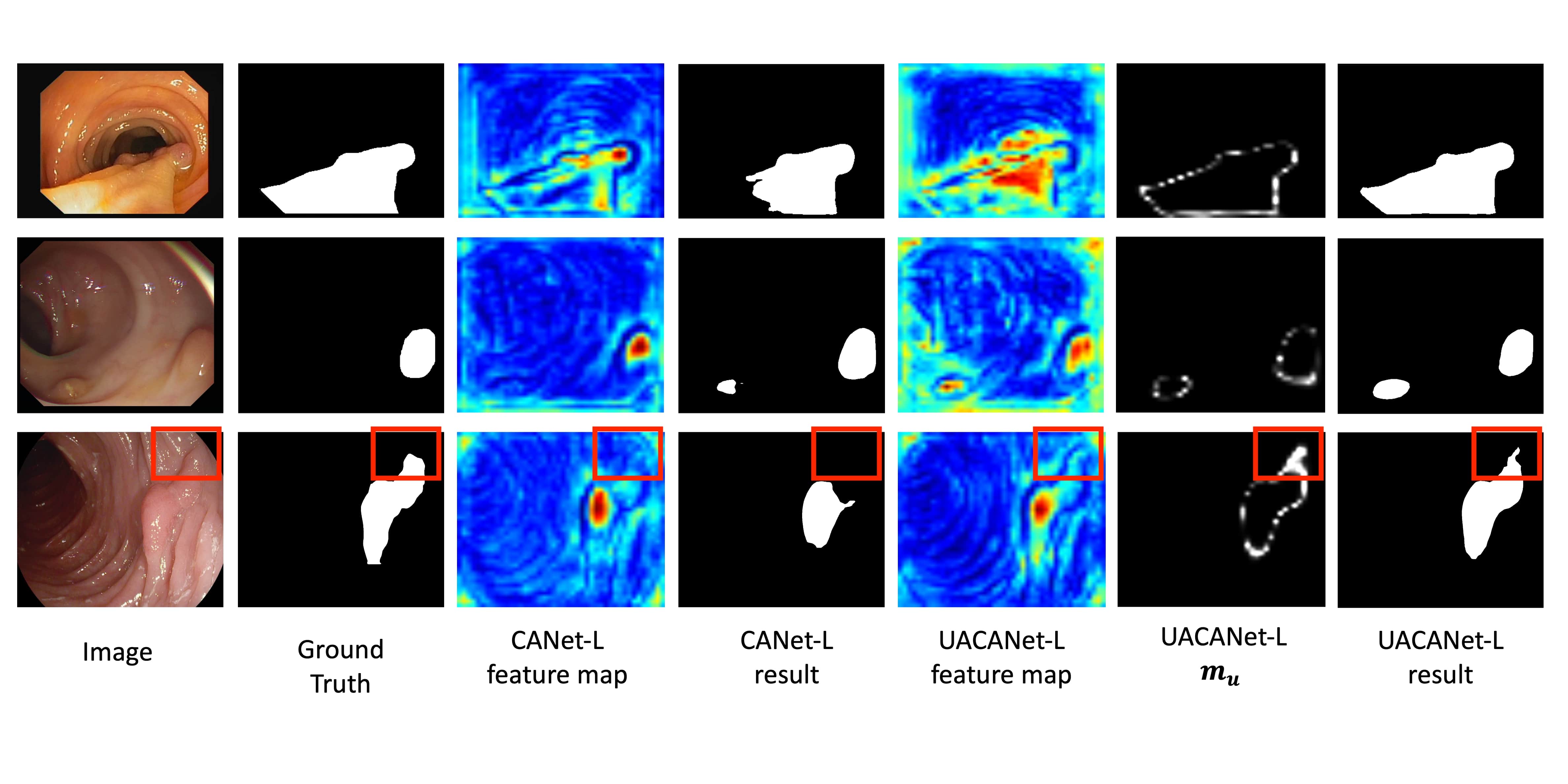}
  \caption{Qualitative results of comparison with CANet-L and UACANet-L}
    \label{fig:5}
\end{figure*}

\begin{equation}
    \label{eq:5}
    \begin{gathered}
    \textbf{t}_i = \delta(s_{fi}  \omega(\textbf{v}_f) + s_{bi}  \omega(\textbf{v}_b) + s_{ui}  \omega(\textbf{v}_u)).
    \end{gathered}
\end{equation}

Note that $\psi(\cdot)$, $\phi(\cdot)$, $\omega(\cdot)$ and $\delta(\cdot)$ are point-wise convolution. Each pixel in context feature map, $\textbf{t}_i$, can be interpreted as a weighted average of three representation vectors $\textbf{v}_f$, $\textbf{v}_b$ and $\textbf{v}_u$. The context feature map $\textbf{t}$ and input feature map $\textbf{x}$ are concatenated with respect to the channel axis and feed forward to the point-wise convolution for final output feature map as shown in \figureautorefname~\ref{fig:4}.

\begin{table}[]
    \begin{center}
        \resizebox{\linewidth}{!}{
        \begin{tabular}{ll|ccc}
            \thickhline
            Dataset & Method & Mean Dice $\uparrow$ & Mean IoU $\uparrow$ & MAE $\downarrow$ \\ \hline \hline
            \multirow{7}{*}{Kvasir} & U-Net \cite{ronneberger2015u} & 0.818 & 0.746 & 0.055 \\
            & U-Net++ \cite{zhou2018unet++} & 0.821 & 0.743 & 0.048 \\
            & ResUNet++ \cite{jha2019resunet++} & 0.813 & 0.793 & - \\
            & SFA \cite{10.1007/978-3-030-32239-7_34} & 0.723 & 0.611 & 0.075 \\
            & PraNet \cite{fan2020pranet} & 0.898 & 0.840 & 0.030 \\ \cline{2-5}
            & \textbf{UACANet-S (Ours)} & \textbf{\textcolor{blue}{0.905}} & \textbf{\textcolor{blue}{0.852}} & \textbf{\textcolor{blue}{0.026}} \\
            & \textbf{UACANet-L (Ours)} & \textbf{\textcolor{red}{0.912}} & \textbf{\textcolor{red}{0.859}} & \textbf{\textcolor{red}{0.025}} \\ \thickhline
            \multirow{7}{*}{\makecell{CVC\\-ClinicDB}} & U-Net \cite{ronneberger2015u} & 0.823  & 0.755  & 0.019 \\
            & U-Net++ \cite{zhou2018unet++} & 0.794 & 0.729 & 0.022 \\
            & ResUNet++ \cite{jha2019resunet++} & 0.796 & 0.796 & - \\
            & SFA \cite{10.1007/978-3-030-32239-7_34} & 0.700  & 0.607  & 0.042 \\
            & PraNet \cite{fan2020pranet} & 0.899  & 0.849  & 0.009 \\ \cline{2-5}
            & \textbf{UACANet-S (Ours)} & \textbf{\textcolor{blue}{0.916}} & \textbf{\textcolor{blue}{0.870}} & \textbf{\textcolor{blue}{0.008}} \\
            & \textbf{UACANet-L (Ours)} & \textbf{\textcolor{red}{0.926}} & \textbf{\textcolor{red}{0.880}} & \textbf{\textcolor{red}{0.006}} \\ \thickhline
        \end{tabular}}
    \end{center}
    \caption{Comparison to the previous \textit{state-of-the-art} methods and our UACANet on Kvasir and CVC-ClinicDB datasets. Red color denotes the best score among the methods, and blue color denotes the second best. $\uparrow$ denotes higher the better and $\downarrow$ denotes lower the better.}
    \label{tab:3}
\end{table}

\section{Experimental Results}

In this section, we demonstrate our implementation details, datasets and benchmarks for experiments, and some experimental results including ablation study on uncertain area and comparison with previous \textit{state-of-the-art} methods on five polyp segmentation benchmarks. We also visualize feature maps and the uncertainty map from UACA and some qualitative results to compare other methods. 

\begin{table*}[]
    \begin{center}
        \resizebox{\textwidth}{!}{
        \begin{tabular}{l|ccc|ccc|ccc}
            \thickhline
            \multirow{2}{*}{\textbf{Method}} & \multicolumn{3}{c}{ETIS} & \multicolumn{3}{c}{CVC-ColonDB} & \multicolumn{3}{c}{CVC-300} \\
            & mDice $\uparrow$ & mIoU $\uparrow$ & MAE $\downarrow$ & mDice $\uparrow$ & mIoU $\uparrow$ & MAE $\downarrow$ & mDice $\uparrow$ & mIoU $\uparrow$ & MAE $\downarrow$ \\ \hline \hline
            U-Net \cite{ronneberger2015u} & 0.398 & 0.335 & 0.036 & 0.512 & 0.444 & 0.061 & 0.710 & 0.627 & 0.022 \\
            U-Net++ \cite{zhou2018unet++} & 0.401 & 0.344 & 0.035 & 0.483 & 0.410 & 0.064 & 0.707 & 0.624 & 0.018 \\
            SFA \cite{10.1007/978-3-030-32239-7_34} & 0.297 & 0.217 & 0.109 & 0.469 & 0.347 & 0.094 & 0.467 & 0.329 & 0.065 \\
            PraNet \cite{fan2020pranet} & 0.628 & 0.567 & 0.031 & 0.709 & 0.640  & 0.045 & 0.871 & 0.797 & 0.010 \\ \hline
            \textbf{UACANet-S (Ours)} & \textbf{\textcolor{blue}{0.694}} & \textbf{\textcolor{blue}{0.615}} & \textbf{\textcolor{blue}{0.023}} & \textbf{\textcolor{red}{0.783}} & \textbf{\textcolor{red}{0.704}} & \textbf{\textcolor{red}{0.034}} & \textbf{\textcolor{blue}{0.902}} & \textbf{\textcolor{blue}{0.837}} & \textbf{\textcolor{blue}{0.006}}\\
            \textbf{UACANet-L (Ours)} & \textbf{\textcolor{red}{0.766}} & \textbf{\textcolor{red}{0.689}} & \textbf{\textcolor{red}{0.012}} & \textbf{\textcolor{blue}{0.751}} & \textbf{\textcolor{blue}{0.678}} & \textbf{\textcolor{blue}{0.039}} & \textbf{\textcolor{red}{0.910}} & \textbf{\textcolor{red}{0.849}} & \textbf{\textcolor{red}{0.005}}\\
            \thickhline
        \end{tabular}}
    \end{center}
    \caption{Comparison to the previous \textit{state-of-the-art} methods and our UACANet on ETIS, CVC-ColonDB and CVC-300 datasets. Red color denotes the best score among the methods, and blue color denotes the second best. $\uparrow$ denotes higher the better and $\downarrow$ denotes lower the better.}
    \label{tab:4}
\end{table*}

\subsection{Implementation Details}

We describe most of our model architecture description in Section \ref{methods} and the number of channels in convolution layers which appear outside of the backbone network is unified as 32 for small model and 256 for large model. We denote the small model with 32 channels as UACANet-S and the large model with 256 channels as UACANet-L. We use Res2Net \cite{gao2019res2net} with $26w \times 4s$ settings as a backbone network. Intermediate backbone feature maps are acquired from each stage's last residual block (green box in \figureautorefname~\ref{fig:1}. For UACANet-L, similar to the DeeplabV3+ \cite{chen2017rethinking}, we modified strides and dilation rates to increase the spatial size of the feature map. We resize images to $352 \times 352$ for both training and inference and resize back to its original size. Unlike PraNet \cite{fan2020pranet} and other \textit{state-of-the-art} methods, we adopt additional data augmentation techniques which are fairly popular in semantic segmentation including random flipping on both horizontal and vertical axis, random image scaling from $0.75$ to $1.25$. We conduct random rotation from $0$ to $359$ degrees since colonoscopy images may rotate during examination. We also add additional random dilation and erosion for ground truth label to enhance generalization. We use Adam optimizer \cite{kingma2014adam} and the initial learning rate is set to $10^{-4}$ with polynomial learning rate decay \cite{chen2017rethinking} with factor $(1-(\frac{iter}{iter_{max}})^{0.9})$. Compared to PraNet which trained only 60 epochs, we increase training epoch to 240 since we apply diverse data augmentation. We use Pytorch \cite{paszke2019pytorch} to implement our model and single Titan RTX GPU for train the model.

\subsection{Datasets}

Following \cite{fan2020pranet}, images which have been randomly selected from Kvasir and CVC-ClinicDB is used for training, but we use same training data for fair comparison which has already been extracted from Kvasir and CVC-ClinicDB and it contains 1450 images total. For benchmark dataset, we use five different datasets.

\begin{description}[leftmargin=0cm]
    \item[CVC-ClinicDB] CVC-ClinicDB \cite{BERNAL201599}, also known as CVC-612 contains 612 images from 25 colonoscopy videos and selected 29 sequences from them. The size of images is $384 \times 288$. 62 images from this dataset are used for test and remaining images are used for training.
    \item[CVC-300] CVC-300 is a test dataset from EndoScene \cite{vazquez2017benchmark}. EndoScene contains 912 images from 44 colonoscopy sequences which were acquired from 36 patients total. Since EndoScene dataset is a combination of CVC-ClinicDB and CVC-300, following D.-P. Fan \textit{et al}, we use CVC-300 as a test dataset which are 60 samples total.
    \item[CVC-ColonDB] CVC-ColonDB \cite{BERNAL20123166} dataset is collected from 15 different colonoscopy sequences and sampled 380 images from these sequences. 
    \item[ETIS] ETIS \cite{Silva_2013} dataset contains 196 images which are collected from 34 colonoscopy videos. The size of images is $1225 \times 966$ which is the largest among other datasets. Unless polyps in this dataset are vary in size and shape, they are mostly small and hard to find, which makes this dataset more challenging.
    \item[Kvasir] Kvasir \cite{jha2020kvasir} Kvasir dataset consists of 1000 polyp images and corresponding annotations. Unlike the other datasets, images vary in size, from $332 \times 487$ to $1920 \times 1072$ and also the size of polyps which appear in the images vary in its size and shape. There are 700 large polyps which is larger than $160 \times 160$, 48 small polyps smaller than $64 \times 64$ and 323 medium polyps within the large and small scale. 900 images are used for training and 100 images are used for test.
\end{description}

\subsection{Ablation Study on Parallel Axial Attention}
\label{ablation paa}
To exhibit the validity of PAA module, we conduct an experiment to evaluate UACANet without PAA modules. We design another model whose specific model architecture is identical to the UACANet except PAA modules (yellow box in \figureautorefname~\ref{fig:3}) is excluded. We choose three metrics to evaluate our methods, mean Dice (mDice), mean intersection over union (mIoU) and mean absolute error (MAE). We choose these two datasets for ablation study since CVC-ClinicDB is sampled for training while ETIS isn't. As shown in \tableautorefname~\ref{tab:1}, UACANet-S with PAA module shows better.

\subsection{Ablation Study on uncertain area}

We conduct an experiment to demonstrate the effectiveness of UACA. We leave all the details identical to the UACA except for excluding the uncertainty map ($\textbf{m}_u$ in \figureautorefname~\ref{fig:4}), namely Context Attention (CA). We substitute UACA with CA in \figureautorefname~\ref{fig:1} to make CANet. We design CANet with same small and large version, namely CANet-S and CANet-L respectively. \tableautorefname~\ref{tab:2} shows quantitative results on CANet and UACANet on CVC-ClinicDB and ETIS. We also choose these two datasets for same reason as \sectionautorefname~\ref{ablation paa}. In terms of performance measure, UACANet consistently outperforms CANet on three major metrics. 

\begin{figure*}[]
  \centering
  \includegraphics[width=\textwidth]{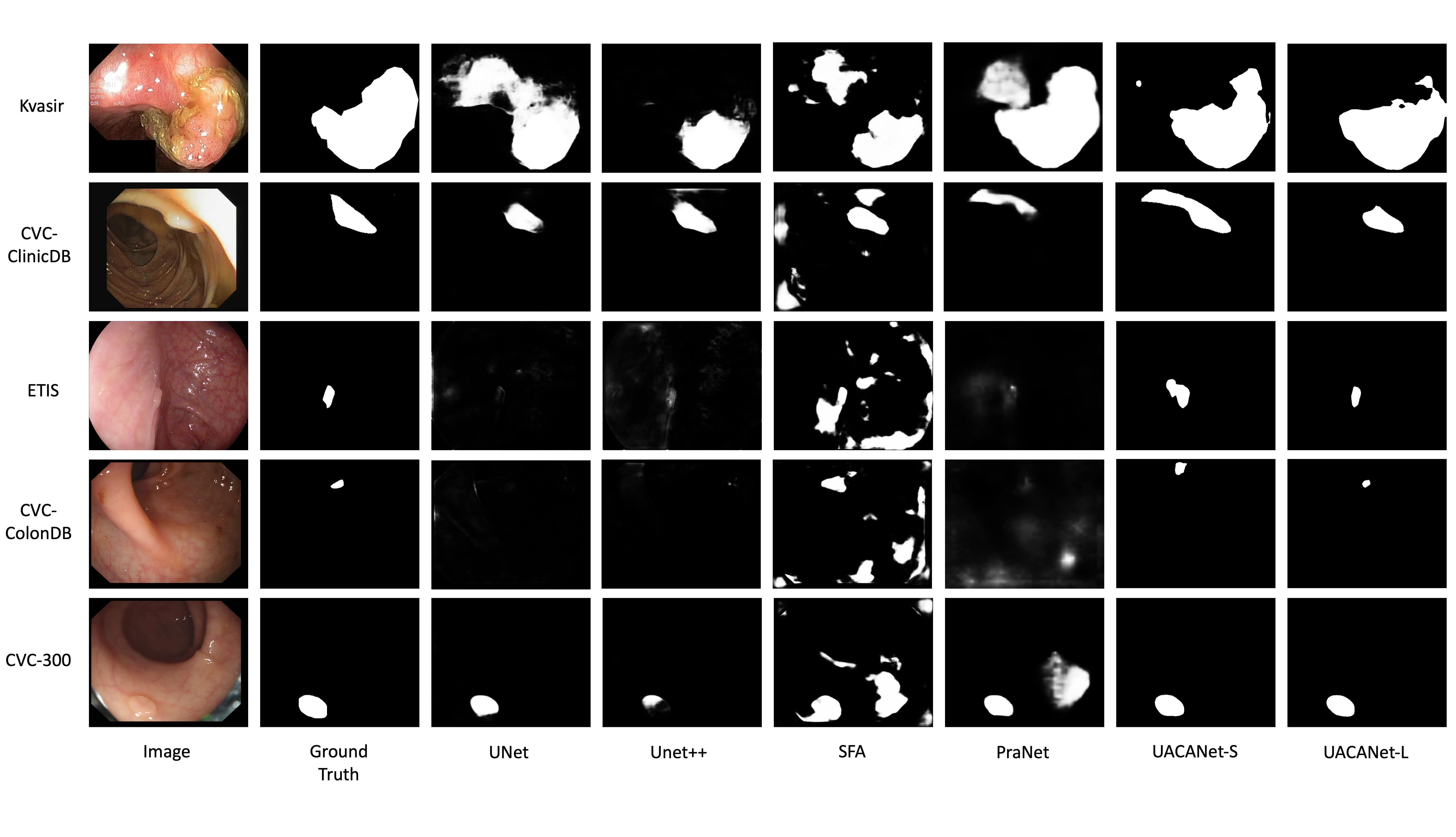}
  \caption{Qualitative results comparison with \textit{state-of-the-art} methods on five different benchmarks.}
    \label{fig:6}
\end{figure*}

We also visualize the output feature map of attention modules in both settings in \figureautorefname~\ref{fig:5} to verify the effectiveness of \textbf{\textit{uncertain area}} qualitatively. UACANet substantially produce more precise results than CANet in terms of quality as well. In first and third row, while visualized feature map and the output of CANet-L shows that even though it detects substantial region of polyps, it fails to detect the ambiguous region which may seems hard to discriminate from mucosa, surface of colon. On the other hand, UACANet consistently segment polyp regions precisely. We also visualize \textbf{\textit{uncertain area}} of UACA ($\textbf{m}_u$ in \figureautorefname~\ref{fig:4}), it is easy to recognize that uncertain area is closely related to the boundary of polyps. Especially for the third row, $\textbf{m}_u$ also helps to detect the ambiguous area denoted as red boxes. Also, in second row, since ground truth miss another polyp on left bottom corner, UACANet detected the missing polyp. 

\subsection{Experiments with State-of-the-Art methods}

As mentioned above, we compare our method with previous \textit{state-of-the-art} methods on five different polyp segmentation benchmarks. Since we train on sampled data from Kvasir and CVC-ClinicDB, even the test results are still unseen, the domain of images from these two datasets is still similar to the training data. Thus, we first demonstrate our results on first two datasets on \tableautorefname~\ref{tab:3}. On both datasets, our UACANet achieve the best performance among other methods. Especially, our UACANet-L achieves 92.6\% mean Dice on CVC-Clinic DB which is 2.7\% improvement over PraNet, the latest \textit{state-of-the-art} method. In \tableautorefname~\ref{tab:4}, we evaluate our method with three completely unseen datasets. We mentioned above that ETIS is the most challenging datasets among other four, nevertheless UACANet-L achieve 76.6\% mean Dice which is 13.8\% improvement over PraNet.

We also demonstrate qualitative results on five benchmarks of previous \textit{state-of-the-art} methods and our method (\figureautorefname~\ref{fig:5}). On Kvasir and CVC-ClinicDB (first and second row), since two datasets are similar to the training dataset, all methods are able to segment the location of polyps, but our method show the most similar results compared to the ground truth. On ETIS dataset (third row), which is the most challenging dataset among other four benchmarks, both UACANet-S and UACANet-L are able to detect a small polyp even if the size of the polyp is very small and hard to notice while other methods has failed to detect.

\section{Conclusion}

We propose a novel polyp segmentation network called UACANet which augments \textbf{\textit{uncertain area}} to the context representation for accurate polyp detection. Without expensive edge annotations, we show that uncertain area is capable of representing boundary information. We propose Parallel Axial Attention for encoder for backbone features and decoder for initial saliency map. We also propose Uncertainty Augmented Context Attention which augments uncertainty area which represents complementary edge information. In a series of both quantitative and qualitative experiments shows that our method outperforms compared to the previous \textit{state-of-the-art} methods.

\section*{Acknowledgement}

This work was supported by Institute of Information \& communications Technology Planning \& Evaluation(IITP) grant funded by the Korea government(MSIT) (No.B0101-15-0266, Development of High Performance Visual BigData Discovery Platform for Large-Scale Realtime Data Analysis), (No.2017-0-00897, Development of Object Detection and Recognition for Intelligent Vehicles) and (No.2018-0-01290, Development of an Open Dataset and Cognitive Processing Technology for the Recognition of Features Derived From Unstructured Human Motions Used in Self-driving Cars)

\bibliographystyle{ACM-Reference-Format}
\balance
\bibliography{ref}

\end{document}